\documentclass[letterpaper]{article} 
\usepackage{aaai23}  
\usepackage{times}  
\usepackage{helvet}  
\usepackage{courier}  
\usepackage[hyphens]{url}  
\usepackage{graphicx} 
\usepackage{natbib}  
\usepackage{caption} 
\usepackage{algorithm}
\usepackage{algorithmic}
\usepackage{bm}
\usepackage{amsmath}
\usepackage{bbding}
\usepackage{newfloat}
\usepackage{listings}
\usepackage{color}

\DeclareCaptionStyle{ruled}{labelfont=normalfont,labelsep=colon,strut=off} 
\floatstyle{ruled}
\newfloat{listing}{tb}{lst}{}
\floatname{listing}{Listing}
\title{Boosting Semi-Supervised Semantic Segmentation with Probabilistic Representations}
\author{
    Haoyu Xie\textsuperscript{\rm 1},
    Changqi Wang\textsuperscript{\rm 1},
    Mingkai Zheng\textsuperscript{\rm 2, 3},
    Minjing Dong\textsuperscript{\rm 2},
    Shan You\textsuperscript{\rm 3},\\
    Chong Fu\textsuperscript{\rm 1, 4},
    Chang Xu \textsuperscript{\rm 2}
}
\affiliations{
    \textsuperscript{\rm 1}School of Computer Science and Engineering, Northeastern University, Shenyang, China\\
    \textsuperscript{\rm 2}School of Computer Science, Faculty of Engineer, The University of Sydney, Sydney, Australia\\
    \textsuperscript{\rm 3}SenseTime Research, Beijing, China\\ 
    \textsuperscript{\rm 4}Key Laboratory of Intelligent Computing in Medical Image, Ministry of Education, Northeastern University, China\\ 
    \{xiehaoyu, wangchangqi\}@stumail.neu.edu.cn, mingkaizheng@outlook.com, mdon0736@uni.sydney.edu.au, youshan@sensetime.com, fuchong@mail.neu.edu.cn, c.xu@sydney.edu.au 

}

\usepackage{bibentry}

\begin{document}

\maketitle


\begin{abstract}
Recent breakthroughs in semi-supervised semantic segmentation have been developed through contrastive learning.
In prevalent pixel-wise contrastive learning solutions, the model maps pixels to deterministic representations and regularizes them in the latent space. 
However, there exist inaccurate pseudo-labels which map the ambiguous representations of pixels to the wrong classes due to the limited cognitive ability of the model.
In this paper, we define pixel-wise representations from a new perspective of probability theory and propose a Probabilistic Representation Contrastive Learning (PRCL) framework that improves representation quality by taking its probability into consideration. Through modelling the mapping from pixels to representations as the probability via multivariate Gaussian distributions, we can tune the contribution of the ambiguous representations to tolerate the risk of inaccurate pseudo-labels. Furthermore, we define prototypes in the form of distributions, which indicates the confidence of a class, while the point prototype cannot. Moreover, we propose to regularize the distribution variance to enhance the reliability of representations.
Taking advantage of these benefits, high-quality feature representations can be derived in the latent space, thereby the performance of semantic segmentation can be further improved.
We conduct sufficient experiment to evaluate PRCL on Pascal VOC and CityScapes to demonstrate its superiority.
The code is available at \textcolor{red}{https://github.com/Haoyu-Xie/PRCL}.
\end{abstract}


\section{Introduction}
Semantic segmentation is a pixel-level classification task, i.e. predicting the class of each pixel.
Existing supervised methods rely on large-scale annotated data, which requires high manual-labeling costs.
Semi-supervised learning \cite{CCT, GCT, pseudoseg, FixMatch} takes advantage of unlabeled data and relieves the labor of human annotation.
Some methods use unlabeled data to improve segmentation models via adversarial learning \cite{GCT}, consistency regularization \cite{cotraining}, and self-training \cite{MT}.
Self-training is a typical solution that uses the prediction generated by a model trained on labeled data (called pseudo-label) as ground-truth to train unlabeled data.

Recently, powerful methods based on self-training \cite{Reco, U2PL} additionally introduce a pixel-wise contrastive learning as an auxiliary task to further explore unlabeled data.
Contrastive learning benefits from not only the local context of neighbouring pixels, but also global semantic class relations across the mini-batch even the entire dataset.
They map pixels to representations and regularize them in the latent space in a supervised way, i.e. gather representations belonging to the same class and scatter representations belonging to different classes, where the semantic class information comes from both ground truths and pseudo-labels.
Most of contrastive learning methods is guided by pseudo-labels in semi-supervised settings.
Therefore, the quality of pseudo-labels is critical since inaccurate pseudo-labels lead to assigning representations to wrong classes and cause a disorder in latent space.
Existing efforts try to polish pseudo-labels via either confidence \cite{Reco} or entropy \cite{DMT}. These techniques can improve the quality of pseudo-labels and eliminate inaccurate ones to some extent. However, the inherent noise as well as essential incorrectness in pseudo-labels are rather difficult to be perfectly tackled by existing work. Thus, we propose to alleviate this risk in an opposite way. Specifically, instead of paying more attention to polishing pseudo-labels for inaccuracy elimination, we propose to improve the quality of representations from data via modelling probability, and allow them to perform better under the inaccurate pseudo labels.
\begin{figure}[t]
  \centering
  \includegraphics[width=1.0\linewidth]{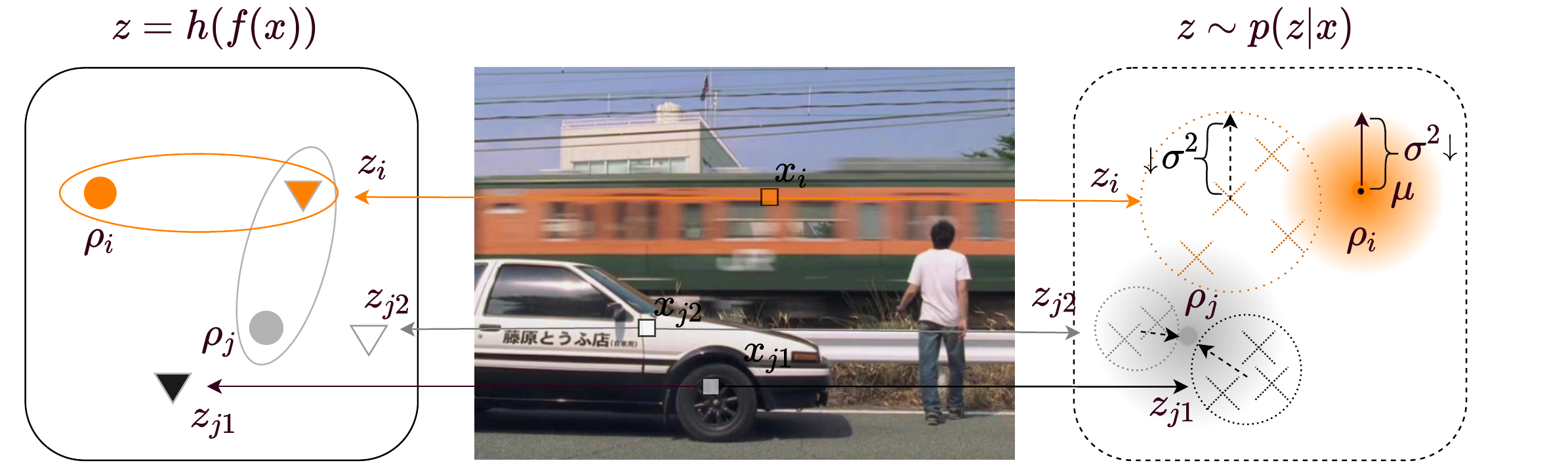}
  \caption{Contradistinction between two types of representations and prototypes. 
  Deterministic representations: left triangle symbol, Probabilistic representations: right dotted cross symbol, Point prototypes: left filled circle symbols, Distribution prototypes: right radial circle symbols.
  }
  \vskip -0.2in
  \label{fig1}
\end{figure}

Comparing with existing conventional deterministic representation modelling, we model representations as a random variable with learnable parameters and represent prototypes in the form of distributions. We take the form of multivariate Gaussian distribution for both representations and prototypes. An illustration of proposed probabilistic representations and distribution prototypes is shown in Figure \ref{fig1}.
The involvement of probability is shown in  $z \sim p(z|x)$. The pixel of the fuzzy train carriage $x_i$ is mapped to $z_i$ in the latent space which contains two parts including the most like representation $\mu$ and the probability $\sigma^2$ which correspond to the mean and variance of distribution respectively. Similarly, the pixels of the car $x_{j1}$ and $x_{j2}$ are mapped to $z_{j1}$ and $z_{j2}$ respectively. For comparison, deterministic mapping is shown in $z=h(f(x))$.
Considering the scenario where the distance from representation $z_i$ to prototype $\rho_i$ is same as the distance from $z_i$ to $\rho_j$, there exist an ambiguity of the mapping from $z_i$ to $\rho_i$ and $\rho_j$ in deterministic representation. On the contrary, $z_i$ is mapped to $\rho_i$ in probabilistic representation since $\rho_i$ has a smaller $\sigma^2$ than $\rho_j$. Note that $\sigma^2$ is inversely proportional to the probability, which implies that the mapping from $z_i$ to $\rho_i$ is more reliable.
Furthermore, $z_{j1}$ and $z_{j2}$ contribute to the car prototype $\rho_j$ to different degrees. Through taking the probability of representations into consideration, the prototypes can be estimated more accurately.
Meanwhile, the variance $\sigma^2$ is constrained during the training procedure, which further improves the reliability of the representations and prototypes.

In this paper, we define pixel-wise representations and prototypes from a new perspective of probability theory and design a new framework for pixel-wise \textbf{P}robabilistic \textbf{R}epresentation \textbf{C}ontrastive \textbf{L}earning, named PRCL.
Our key insight is to: (i) involve modelling probability into the representations and prototypes, and (ii) explore a more accurate similarity measurement between probabilistic representations and prototypes.
For the first objective (i), we concatenate an Probability head (Multilayer Perceptron, MLP) to encoder to predict the probabilities of representations and construct a distribution prototype with probabilistic representations as observations based on Bayesian Estimation \cite{Bayesian}.
In the latent space, each prototype is represented as a distribution rather than a point, which enables them to explore the uncertainty.
For objective (ii), we leverage mutual likelihood score (MLS) \cite{Probabilistic_face_embeddings} to directly compute the similarity among probabilistic representations and distribution prototypes.
MLS can naturally adjust the weight of distance based on the uncertainty, i.e. penalize ambiguous representations and vice versa.
Taking the advantage of the confidence information contained in probabilistic representations, model robustness to inaccurate pseudo-labels is significantly enhaned for stable training.
In addition, we propose a soft freezing strategy to optimize probability head free from probability converging sharply to $\infty$ during training without constraint. 

In summary, we propose to alleviate the negative effects from inaccurate pseudo-labels by introducing probabilistic representation with PRCL framework, which reduces the contribution of representations with high uncertainty and concentrates on more reliable ones in contrastive learning. To the best of our knowledge, we are the first to simultaneously train the representation and probability. Extensive evaluation on Pascal VOC \cite{pascal} and CityScapes \cite{cityscapes} to demonstrate the superior performances than the SOTA baselines.


\section{Related Work}
\subsection{Semi-supervised semantic segmentation}
The goal of semantic segmentation is to classify each pixel in an entire image by class.
The training of such dense prediction tasks relies on large amounts of data and tedious manual annotations.
Semi-supervised learning is a label-efficient task that needs to take advantage of a large amount of unlabeled data to improve model performance.
Entropy minimization \cite{adversarial4ssl, three-stage} and consistency regularization \cite{CCT, cotraining, UCC} are two main branches.
Recently, self-training methods benefit from strong data augmentation \cite{cutmix, classmix, AEL} and well-refined pseudo-labels \cite{FixMatch, DMT}.
Besides, some methods \cite{unbias_subclass} balancing the distributions of subclass are competitive in some scenarios. 
Recent works based on self-training \cite{Reco, U2PL, SePiCo} attempt to regularize representations in latent space for better embedding space distribution.
This improves the quality of features and leads to better model performance, which is also our goal.
\subsection{Contrastive Learning}
As a major branch of metric learning, the key idea of contrastive learning is to pull positive pairs close and push negative pairs apart in the latent feature space through a contrastive loss.
At the instance level, it treats each image as a single class and distinguishes the image from others in multiple views \cite{InstDisc, yemang_InstFeat, SimCLR, MoCo, BYOL}.
To alleviate the negative impact of sampling bias, some works \cite{debiased_contrast} try to correct for the sampling of same-label data, even without the information of true labels.
Furthermore, in some supervised or semi-supervised settings, some works \cite{lassl} introduce class information to train models to distinguish between classes.
At the pixel level, pixel-wise representations are distinguished by labels or pseudo-labels \cite{s4_with_context, cross_image_pixel4ss}.
However, in the semi-supervised setting, only a small amount of labeled data is available.
Most pixel divisions are based on pseudo-labels, and inaccurate pseudo-labels lead to a disorder in the latent space.
To address these issues, previous methods \cite{Reco, s4_with_mb} try to polish pseudo-labels.
In our approach, we focus on tolerating inaccurate pseudo-labels rather than filtering them.

\subsection{Probabilistic Embedding}
Probabilistic Embeddings (PE) is an extension of conventional embeddings.
Methods of PE usually predict the overall distribution of the embeddings, e.g. , Gaussian \cite{Probabilistic_face_embeddings} and von Mises-Fisher \cite{Spherical_confidence_learning_for_face_recognition}, rather than a single vector.
The ability of neural networks to predict distributions stems from the work of Mixture Density Networks (MDN) \cite{MDN}.
Later, Variable Auto-Encoders (VAE) \cite{VAE} first proposed the use of MLP to predict the mean and variance of a distribution.
Most of works on probabilistic embeddings \cite{Probabilistic_face_embeddings, Stochastic_prototype_embeddings, vMF_loss, Probabilistic_representations_for_video_contrastive_learning} are based on this architecture.
Hedged Instance emBeddings (HIB) \cite{HIB} is the first attempt to apply PE to image retrieval and verifications.
Further, PE is applied to face verification tasks.
In Probabilistic Face Embeddings (PFE) \cite{Probabilistic_face_embeddings}, each image is mapped to a Gaussian distribution in the latent space, with the mean produced by a pre-trained model and variance predicted via an MLP (some works call it uncertainty head, we name it probability head in our work).
And Sphere Confidence Face (SCF) \cite{Spherical_confidence_learning_for_face_recognition} maps image to a von Mises-Fisher distribution with mean and concentration parameter.
We borrow the same architecture, but it is worth noting that PFE and SCF optimize mean and variance (concentration parameter) in two stages, i.e. pre-train a deterministic model to predict the mean and freeze it to optimize the variance.
But in our work, the mean and variance are optimized simultaneously, which allows them to interact with each other.

Although the conventional distance metric can measure the similarity of distribution using their mean, it is insufficient to represent the probabilistic similarity due to variance.
To address this issue, HIB uses the reparameter trick \cite{VAE} to obtain two sets of samples from two distributions through Monte-Carlo sampling, and accumulates the similarity of samples to represent the distribution similarity.
PFE and SCF directly compute distribution similarity using mutual likelihood score, but they can not simultaneously optimize mean and variance.
This is because uncertainty/probability (variance) is only valuable if representation (mean) is reasonable.
PFE and SCF achieve this by training the mean and variance in two stages, and we address this by training them separately with different learning rates.
\begin{figure}[t] 
  \centering
  \includegraphics[width=1.0\linewidth]{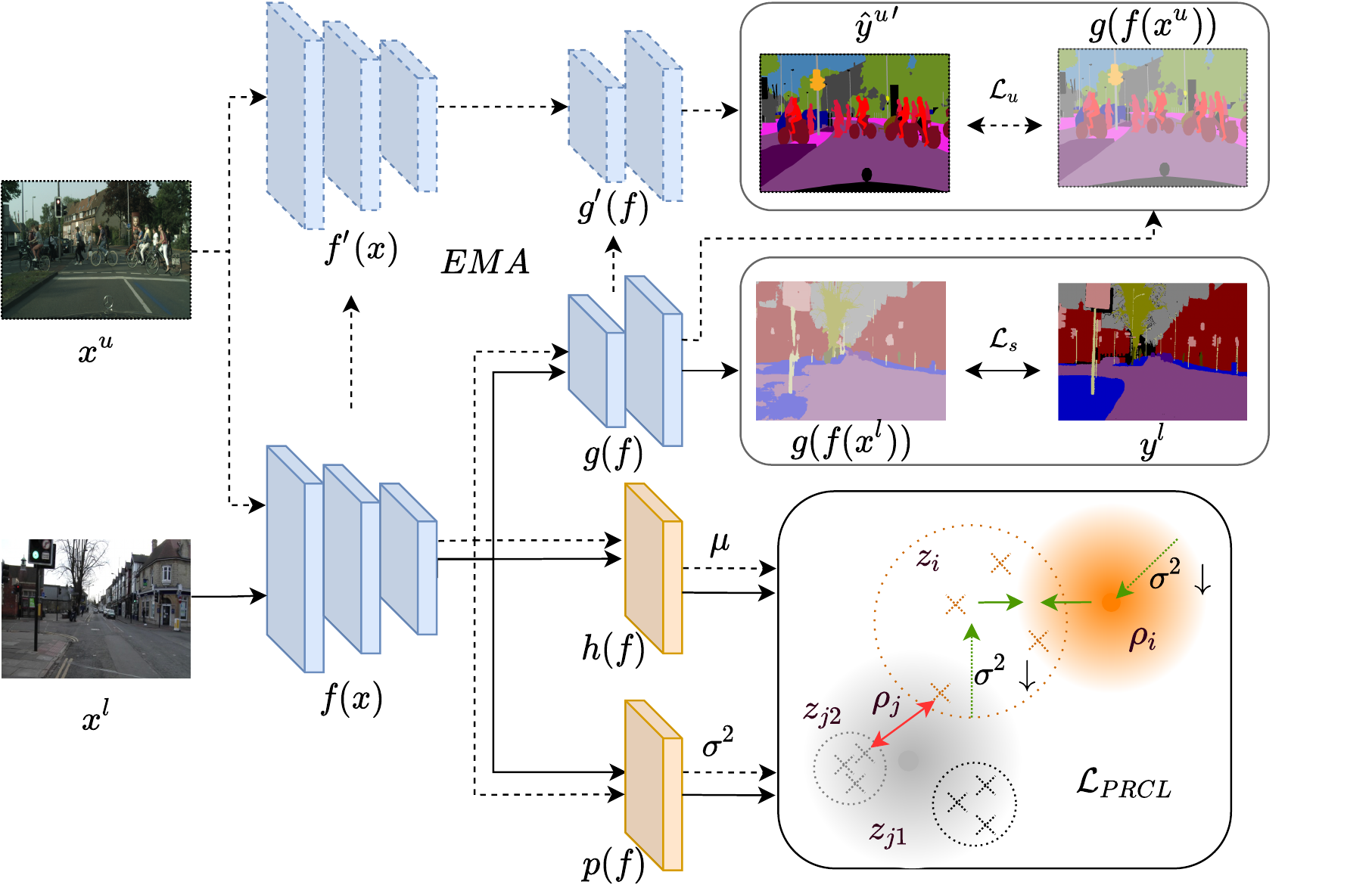}
  \caption{Overview of PRCL applied to MT framework.
  }
  \vspace{-10pt}
  \label{fig2}
\end{figure}
\section{Methodology}

\subsection{Preliminaries}
The objective of semi-supervised semantic segmentation is to train a model with only a limited amount of labeled data $\mathcal{D}_l=\{(\bm{x}^l_i, \bm{y}_i^l)\}_{i=1}^{N_l}$ and substantial unlabeled data $\mathcal{D}_u=\{\bm{x}^u_i\}^{N_u}_{i=1}$. An illustration of our framework is shown in Figure \ref{fig2}. Following the typical self-training frameworks, our framework consists of two networks with the same architecture, named student network and teacher network respectively.
Each model contains an encoder $f(\cdot)$ and a predictor $g(\cdot)$. We denote $f(\cdot)$, $g(\cdot)$ as the encoder and predictor of student network and $f^\prime(\cdot)$, $g^\prime(\cdot)$ as those of teacher network.
The student network parameters are optimized via stochastic gradient descent (SGD) to minimize the loss function $\mathcal{L}$ while the teacher network parameters are updated by Exponential Moving Average (EMA) of the student network parameters.
In addition, in contrastive learning-based semi-supervised methods, a representation head $h(\cdot)$ is introduced to student network to map pixels to representations and a contrastive loss $\mathcal{L}_{c}$ is computed between them.
For simplicity, we denote the one hot coding format of the pseudo-label from teacher network $g^\prime(f^\prime(\bm{x}^u))$ as $\hat{\bm{y}}^{u\prime}$ and the projection result $h(f(\bm{x}))$ as $\bm{z}$.
A supervised cross entropy loss $\mathcal{L}_{s}$ is calculated between the prediction of student network $f(g(\bm{x}^l))$ and ground truth $\bm{y}^l$.
And an unsupervised cross entropy loss $\mathcal{L}_{u}$ is calculated between the prediction of the student network $g(f(\bm{x}^u))$ and the pseudo-label $\hat{\bm{y}}^{u\prime}$ from teacher network.
The overall loss $\mathcal{L}$ for contrastive learning-based semi-supervised segmentation is formulated as
\begin{equation}\label{eq1}
    \scalebox{0.9}{$ \displaystyle
            \mathcal{L} = \mathcal{L}_{s} + \lambda_u \mathcal{L}_{u} + \lambda_c \mathcal{L}_{c},
    $}
\end{equation} 
\begin{equation}
    \scalebox{0.9}{$ \displaystyle
    \mathcal{L}_s = \frac{1}{\lvert \mathcal{B}_l \rvert } \sum_{({\bm{x}}^l_i,\bm{y}_i^l)\in \mathcal{B}_l} \ell_{ce}(g(f(\bm{x}_i^l)),\bm{y}_i^l),
    $}
    \nonumber
\end{equation}
\begin{equation}
    \scalebox{0.9}{$ \displaystyle
    \mathcal{L}_u = \frac{1}{\lvert \mathcal{B}_u \rvert } \sum_{\bm{x}_i^u\in \mathcal{B}_u} \ell_{ce}(g(f(\bm{x}_i^u)), \hat{\bm{y}}^{u\prime}_i),
    $}
    \nonumber
\end{equation}
where $\lambda_u$ and $\lambda_c$ are two hyper-parameters to adjust the contributions of $\mathcal{L}_u$ and $\mathcal{L}_c$ respectively, and $\mathcal{B}_l$ and $\mathcal{B}_u$ denote labeled images and unlabeled images in a mini-batch respectively.
The contrastve learning loss $\mathcal{L}_c$ is computed by infoNCE \cite{CPC} as follows:
\begin{equation}\label{eq2}
    \scalebox{0.9}{$ \displaystyle
    \begin{aligned}
        \mathcal{L}_c=& -\frac{1}{|C| \times |\mathcal{Z}_c|} \sum_{c \in C}\sum_{\bm{z}_{ci} \in \mathcal{Z}_c}\\
        &log[\frac{e^{s(\bm{z}_{ci},\bm{\rho}_c)/ \tau}}
        {e^{s(\bm{z}_{ci}, \bm{\rho}_c))/ \tau}+\sum_{\tilde{c} \in \tilde{C}} \sum_{\bm{z}_{\tilde{c} i} \in \mathcal{Z}_{\tilde{c}}}e^{s(\bm{z}_{ci},\bm{z}_{\tilde{c}i})/ \tau}}],
    \end{aligned}
    $}
\end{equation}
where $C$ denotes the set including all available classes in a mini-batch, $\mathcal{Z}_c$ denotes the set including sampled anchor representations belonging to anchor class $c$, $\tilde{C}$ denotes the set of classes other than $c$, $\mathcal{Z}_{\tilde{c}}$ denotes the set including sampled negative representations belonging to class $\tilde{c}$, $\tau$ denotes the temperature control of the softness of the distribution, and $s$ denotes similarity measurement, e.g., cosine similarity and $\ell_2$ distance, in our work, mutual likelihood score (MLS) \cite{Probabilistic_face_embeddings} is leveraged.
In addition, $\bm{z}_c$ are representations mapped from the pixels belonging to the same class $c$, and the class information is obtained from labels $\bm{y}^l$ of labeled data and pseudo-labels $\hat{\bm{y}}^{u\prime}$ of unlabeled data.
The goal of contrastive learning is to improve the representation capability of the encoder $f(\cdot)$ by regularizing the representations for the better performance in the downstream semantic segmentation tasks.
Considering an ideal representation space $\mathcal{Z}$ for downstream tasks, the similar representations $\bm{z}$ of the same class $c$ are concentrated around the corresponding prototype $\bm{\rho}_c$, while the prototypes of different classes should be separated from each other.
However, due to inaccurate pseudo-labels, there is a mismatch between $\bm{z}_c$ and $c$.
Our goal is to further improve the representation performance for the robustness under this case.

\subsection{Probabilistic Representation}
The representations in conventional contrastive learning are deterministic.
%
However, there exist inaccurate pseudo-labels which map the ambiguous representations of pixels to the wrong classes due to the limited cognitive ability of the model.
Concretely, some pixels from the class $C$ will be treated as one of another classes $\tilde{C}$, leading to a perturbation in $\mathcal{L}_c$ calculating.
Although it is difficult to perform complete elimination of inaccurate pseudo-labels, we can model the mapping probability to measure its confidence, thereby reducing low confidence mapping caused by inaccurate pseudo-labels.
We denote the probability of mapping a pixel $\bm{x}_i$ to a representation $\bm{z}_i$ as $p(\bm{z}_i|\bm{x}_i)$ and define the representation as a random variable following it. 
For simplicity, we take the form of multivariate Gaussian distribution $\mathcal{N}(\bm{\mu},\bm{\sigma}^2\bm{I})$ as
\begin{equation}\label{eq3}
     p(\bm{z}_i|\bm{x}_i) = \mathcal{N}(\bm{z}; \bm{\mu}_i,\bm{\sigma}^2_i\bm{I}),
\end{equation}
where $\bm{I}$ represents the unit diagonal matrix.
In this form, $\bm{\mu}$ can be viewed as most like representation values, and $\bm{\sigma}^2$ can show the probability in the representation values.
It is worth noting that $\bm{\sigma}^2$ is inversely proportional to probability, i.e. the greater the $\bm{\sigma}^2$, the lower the probability, which is consistent with the form in MLS.
Both dimensions of $\bm{\mu}$ and $\bm{\sigma}^2$ are the same.
The mean $\bm{\mu}$ is predicted by the representation head $h(\cdot)$. Meanwhile, we introduce a probability head $p(\cdot)$ in parallel to predict the variance $\bm{\sigma}^2$, as shown in Figure \ref{fig2}.
To achieve the better representation learning for downstream semantic segmentation tasks, the same class representations should concentrate on their prototypes in the latent space.
The prototype can be calculated with the mean of the same class representations in deterministic representation methods as
\begin{equation}\label{eq4}
    \bm{\rho} = \frac{1}{n}\sum_{i=1}^n \bm{z_i},
\end{equation}
where $n$ is the number of representations sampled.
This method has a limitation that all representations contribute the same to the prototype.
In our method, instead, we estimate it using Bayesian Estimation incorporating new probabilistic representations as observations.
With probabilistic representations, a conjugate formula can be derived for prototype estimation.
The prototype is the posterior distribution after the $n^{th}$ observations $\{\bm{z}_1, \bm{z}_2, ..., \bm{z}_n\}$.
Under the assumption that all the observations are conditionally independent, the distribution prototype can be derived as
\begin{equation}\label{eq5}
    \scalebox{0.92}{$ \displaystyle
    p(\bm{\rho}|\bm{z}_1, \bm{z}_2, ..., \bm{z}_{n+1})=\alpha\frac{p(\bm{\rho}|\bm{z}_{n+1})}{p(\bm{\rho})}p(\bm{\rho}|\bm{z}_1, \bm{z}_2, ..., \bm{z}_n),
    $}
\end{equation}
where $\alpha$ is a normalization factor.
In addition to Equation \ref{eq3}, we can rewrite the prototype as
\begin{equation}\label{eq6}
    \bm{\rho} \sim \mathcal{N}(\hat{\bm{\mu}}, \hat{\bm{\sigma}}^2\bm{I}), 
\end{equation}
where
\begin{equation}\label{eq7}
    \hat{\bm{\mu}} = \sum_{i=1}^n\frac{\hat{\bm{\sigma}}^2}{\bm{\sigma}_i^2}\bm{\mu}_i
\end{equation}
\begin{equation}\label{eq8}
    \frac{1}{\hat{\bm{\sigma}}^2} = \sum_{i=1}^n\frac{1}{\bm{\sigma}_i^2}.
\end{equation}
From the Equation \ref{eq7}, we can know that the representations with different probabilities contribute differently to the prototype.
The smaller $\bm{\sigma}_i^2$ (higher probability) corresponds to more contribution, and vice versa.
And the negative impact of high-risk false representations will be reduced, so the prototypes can be estimated reliably.
And with the $\hat{\bm{\sigma}}^2$, the distribution prototype act as a radius region in the latent space, which can express the potential location of the exact prototype with probability.
As Equation \ref{eq8} shows, during the training, with more representations accumulation, the $\bm{\sigma^2}$ of prototype decreases.
The region of distribution prototype becomes smaller. In other words, the estimated prototype becomes clearer.
Proof details refer to \cite{Probabilistic_face_embeddings}.

\subsection{Probabilistic Representation Contrastive Learning}
According to the previous section, we redefine $\bm{z}$ and $\bm{\rho}$ in Equation \ref{eq2}.
However, conventional $\ell_2$ distance does not have the ability to measure the similarity between two distributions.
To solve this problem, we leverage the Mutual likelihood Score (MLS) to measure the similarity between two distributions $\bm{z}_i$ and $\bm{z}_j$, as follows:
\begin{equation}\label{eq9}
\scalebox{0.92}{$ \displaystyle
    \begin{aligned}
        MLS(\bm{z}_i,\bm{z}_j)=& log(p(\bm{z_i}=\bm{z}_j))\\
        =&-\frac{1}{2}\sum_{l=1}^D(\frac{(\mu_i^{(l)}-\mu_j^{(l)})^2}{\sigma_i^{2(l)}+\sigma_j^{2(l)}} + log(\sigma_i^{2(l)}+\sigma_j^{2(l)}))\\
        &-\frac{D}{2}log2\pi,
    \end{aligned}
    $}
\end{equation}
where $\mu_i^{(l)}$ refers to the $l^{th}$ dimension of $\bm{\mu}_i$ and the same for $\sigma_i^{(l)}$.
MLS is a combination of a weighted $\ell_2$ distance and a log regularization term, essentially.
Conventional $\ell_2$ distance only consider the similarity between representations mapped in the latent space by the pseudo-labels without considering their reliability.
The inaccurate pseudo-labels leads to a wrong optimizing direction, which destructs the latent space.
We argue that inaccurate pseudo-labels often come with the low probabilities.
With probabilities of $\bm{z}_i$ and $\bm{z}_j$, MLS responds to inaccurate pseudo-labels from two perspectives.
\textbf{(\romannumeral1)}: In the first term, the weight of $\ell_2$ distance is small when the $\bm{\sigma}^2$ is large, which indicates that the similarity between $\bm{z}_i$ and $\bm{z}_j$ becomes lower due to the low probabilities, even if they are very similar in the view of $\ell_2$ distance. 
The probability has been taken into consideration besides the simple similarity measure for representations learning.
\textbf{(\romannumeral2)}: In the second term, $\bm{\sigma}^2$ is penalized for the low probability representations, which makes all the representations more reliable.
Besides, $\bm{\sigma}^2$ and $\bm{\mu}$ can interact with each other.
The learnable $\bm{\sigma}^2$ is associated with $\ell_2$ distance.
This means that $\bm{\sigma}^2$ can be learned via the relations among representations.
On the other hand, the $\bm{\mu}$ can also be optimized via the $\bm{\sigma}^2$.
This is consistent with intuitive cognition.


We introduce Equation \ref{eq9} into Equation \ref{eq2} and rewrite it as
\begin{equation}\label{eq10}
\scalebox{0.84}{$ \displaystyle
    \begin{aligned}
        \mathcal{L}_c=&-\frac{1}{\lvert C \rvert\times \lvert \mathcal{Z}_c\rvert}\sum_{c\in C}\sum_{\bm{z}_{ci}\in\mathcal{Z}_c}\\
        &log[\frac{e^{MLS(\bm{z}_{ci},\bm{\rho}_c)/\tau}}
        {e^{MLS(\bm{z}_{ci}, \bm{\rho}_c))/\tau}+\sum_{\tilde{c}\in\tilde{C}} \sum_{\bm{z}_{\tilde{c} i}\in \mathcal{Z}_{\tilde{c}}}e^{MLS(\bm{z}_{ci},\bm{z}_{\tilde{c}i})/\tau}}],
    \end{aligned}
    $}
\end{equation} 
and name it PRCL loss $\mathcal{L}_{PRCL}$, whose pseudo-code is shown in Appendix A.
Like conventional contrastive learning, the negatives push each other away, while the positives converge toward the prototype.
In addition, optimized with the PRCL loss, the $\bm{\sigma}^2$ of probabilistic representations on a downward trend, i.e. the representations get more and more certain.
With the accumulation of representations, the $\bm{\sigma}^2$ of prototypes are decreasing, i.e. the prototype becomes clearer, which is verified in our empirical evaluation of their tendency during training in Experiment Section.
And we apply PRCL loss to both labeled and unlabeled data.
Cooperated with supervised cross-entropy loss $\mathcal{L}_{s}$ and unsupervised cross-entropy loss $\mathcal{L}_{u}$, the overall loss $\mathcal{L}$ for semi-supervised segmentation is rewrited as
\begin{equation}\label{eq11}
    \mathcal{L} = \mathcal{L}_{s} + \lambda_u \cdot \mathcal{L}_{u} + \lambda_c(t) \cdot \mathcal{L}_{PRCL}.
\end{equation}
In $\mathcal{L}_{u}$, we only consider the pixels whose predicted confidence $\hat{\bm{y}}_q$ (the maximum of prediction after SoftMax operation) are greater than $\delta_{u}$, and $\lambda_u$ is defined as the percentage of them, following the prior method \cite{classmix}.
$\lambda_c(t)$ is a time-variant scaling parameter to control the weight of $\mathcal{L}_{PRCL}$. We refer more details to the Experiment Section.

\subsection{Soft Freezing}
We find optimization instability where $\bm{\sigma}^2$ would converge to $\infty$ if left unbounded.
At the beginning of training, the probability head output increases substantially, causing training to crash.
We mainly attribute it to the fact that the representations is unreasonable at the beginning of the training.
At this stage, optimizing probability with unreasonable representations is meaningless.
It is crucial to use some empirical tricks to avoid it.
Some works \cite{Probabilistic_face_embeddings, Spherical_confidence_learning_for_face_recognition} freeze the trained representation head when training probability head.
Some works pre-train the representations and freeze it when training probability head freezing the trained representations later.
In addition, an additional KL regularization term between the distribution and the unit Gaussian prior $\mathcal{N}(0,I)$ is employed to prevent the $\bm{\sigma}^2$ from converging to $\infty$.
However, intuitively, this trick will hinder the interaction between probability and representation, thus limiting the performance of the probability.
To address this issue, we separate the training of the probability head from the ensemble and train it with a much smaller learning rate in our work.
This makes probability head training keep pace with others for stable optimization and enables simultaneous training of the probability and the representation to interact with each other.
We name this empirical trick Soft Freezing.
\section{Experiments}
\textbf{Dataset.} We conduct experiments on Pascal VOC 2012 \cite{pascal} dataset and CityScapes \cite{cityscapes} dataset to test the effectiveness of PRCL.
The Pascal VOC 2012 contains 1464 well-annotated images for training and 1449 images for validation originally. Following \cite{Reco, U2PL, ST++}, we introduce SBD \cite{SBD} as additional training data into training set.
CityScapes is an urban scene dataset which includes 2975 training images and 500 validation images.

\noindent\textbf{Network structure.} We use DeepLabv3+ \cite{DeepLab} with ResNet-101 \cite{resnet} pre-trained on ImageNet \cite{imagenet}.
The prediction head and representation head are composed of \texttt{Conv-BN-ReLU-Conv}.
The probability head is composed of \texttt{Linear-BN-ReLU-Linear-BN}.

\noindent\textbf{Sampling strategy.} Due to limited computation and memory, it would be unachievable for us to sample all pixels in the training set. In our method, we follow the prior work \cite{Reco} and adopt some sampling strategies. 
(a) Valid samples sampling strategy: In order to artificially avoid some noise and refine more meaningful representations, we set a threshold $\delta_w$ for sampling valid representations.
Representations will be valid only if its corresponding $\hat{\bm{y}}_q$ is higher than $\delta_w$.
(b) Anchor sampling strategy: For the purpose of paying more attention to relatively ambiguous representations, we set a threshold $\delta_s$ for $\hat{\bm{y}}_q$ and randomly sample a suitable number of hard anchors whose $\hat{\bm{y}}_q$ are below $\delta_s$, which makes our training focus on the ambiguous representations.
(c) Negative samples sampling strategy: We non-uniformly sample a suitable number of negative representations $\bm{z}_{\tilde{c}}$ for each anchor.
The sampling ratio is based on the similarity between prototype for negative representation classes $\bm{\rho}_{\tilde{c}}$ and that for current anchor class $\bm{\rho}_c$.

\noindent\textbf{Experimental details.} We adjust PRCL contribution with a loss scheduler.
Mathematically, given the total training epochs $T_{total}$ and the initial weight $\lambda_c(0)$, the weight $\lambda_c$ at the $t$-th epoch can be calculated as,
\begin{equation}\label{eq12}
    \lambda_c(t) = \lambda_c(0) \cdot exp(\alpha \cdot (\frac{t}{T_{total}})^2)
\end{equation}
where $\alpha$ is a negative constant, which determines the rate of weight descent.

\noindent\textbf{Evaluation.} We adopt the mean of Intersection over Union (mIoU) as the metric to evaluate the performance of our work. All results are measured on the \texttt{val} set in both Pascal VOC and CityScapes.

\subsection{Probability Behavior}
\begin{figure}[t]
  \centering
  \includegraphics[width=0.9\linewidth]{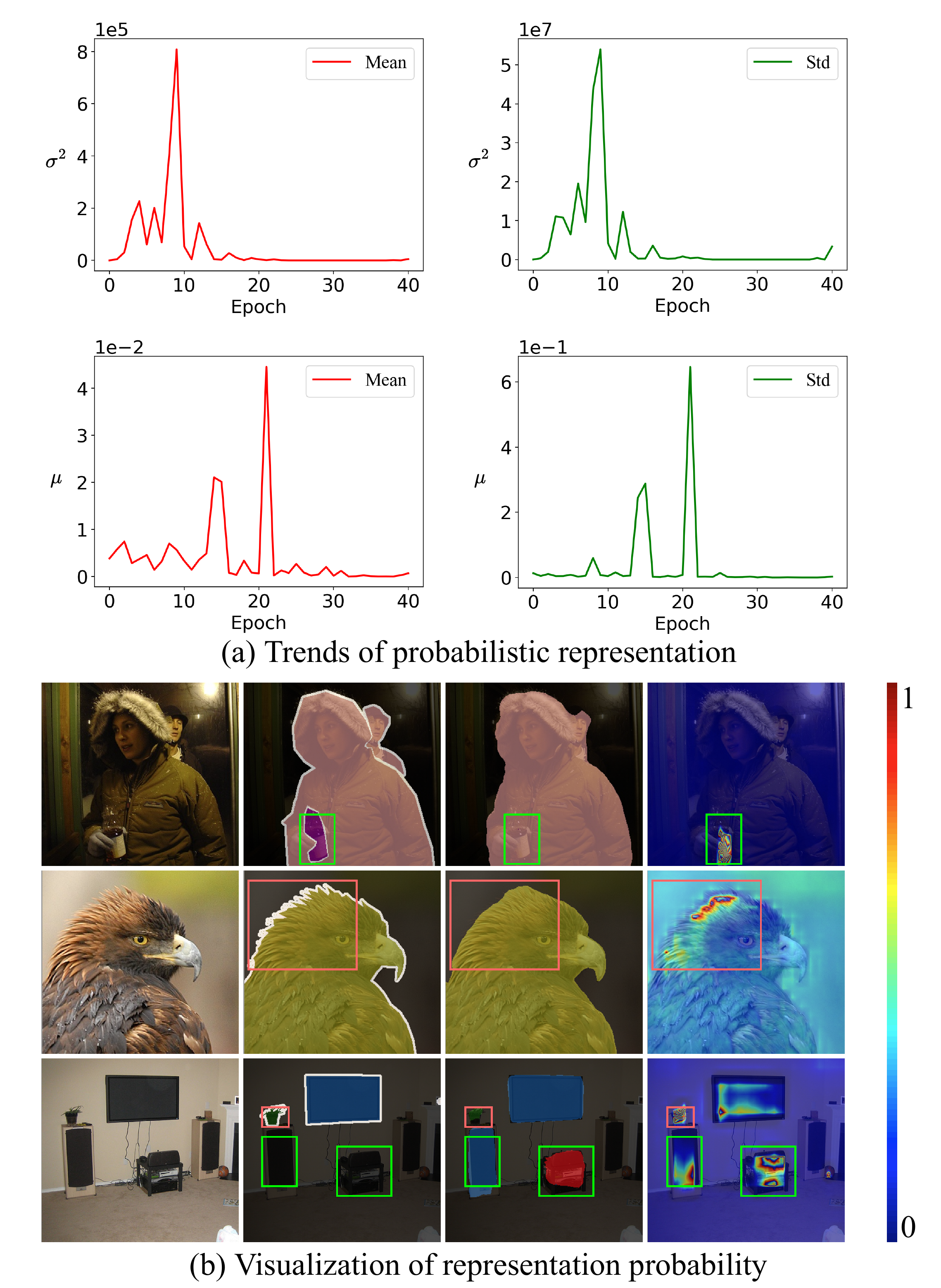}
  \vspace{-5pt}

  \caption{(a) Value of representation probability. (b) Visualization of representation probability.
   }
   \vspace{-8pt}
  \label{fig3}
\end{figure}

Figure \ref{fig3} shows the behavior of the probability (The $\ell_1$ norm of $\bm{\sigma}^2$) and reflects the relationship between probability and inaccurate pseudo-labels.
In Figure \ref{fig3}(a), first row represents the mean and standard deviation of $\bm{\sigma}^2$ in representation during training, and second row represents the same statistics of $\bm{\mu}$.
At the beginning of training process, representations are unreasonable, which lead to a large $\bm{\sigma}^2$.
When representations are more and more certain, $\bm{\sigma}^2$ is decreasing and is more meaningful.
In Figure \ref{fig3}(b), columns from left to right represent input image, ground-truth, pseudo-label, and probability, respectively. For the fourth column, the red color represents the large $\bm{\sigma}^2$ (low probability).
The green boxes mark the mismatches caused by inaccurate pseudo-labels (e.g., person and bottle) and the red boxes mark the fuzzy pixels (e.g., furry edge of the bird).
These two cases are highlighted by $\bm{\sigma}^2$ and make low contribution in training process.
\subsection{Comparison with Existing Methods}
In this subsection, firstly, we reproduce two semi-supervised segmentation baselines: Mean Teacher (MT) and ClassMix \cite{classmix} on Pascal VOC and CityScapes.
In addition, we also conduct the experiment only on labeled dataset (Supervised) to demonstrate our effective use of abundant unlabeled data.
These methods are conducted with standard DeepLabv3+ and ResNet-101 under the same settings.
Besides, for fair comparing, we use the modified ResNet-101 with the deep stem block as our backbone to compare with state-of-the-art methods, following previous works \cite{CPS, U2PL, ST++}.
In addition, our labeled dataset splits are also derived from the previous work \cite{U2PL, ST++, psmt}.

\begin{table}[t]
\centering
\caption{Results on Pascal VOC. All results are reproduced.}
\vspace{-7pt}
\setlength{\tabcolsep}{2mm}{%
\begin{tabular}{cccc}
\hline
\multicolumn{4}{c}{Pascal VOC}                     \\ \hline
Method           & $46$ labels     & $60$ labels   & $92$ labels \\ \hline
Supervised       & $41.18$         & $48.00$       & $52.38$         \\
MT               & $39.04$         & $44.97$       & $49.11$             \\
ClassMix         & $41.55$         & $53.21$       & $63.26$          \\ \hline
PRCL(w/ClassMix) & $\bm{43.00}$    & $\bm{58.10}$  & $\bm{68.52}$          \\ \hline
\end{tabular}
}
\label{tab1}
\end{table}
\begin{table}[t]
\centering
\caption{Results on CityScapes. All results are reproduced.}
\vspace{-7pt}
\setlength{\tabcolsep}{2mm}{%
\begin{tabular}{cccc}
\hline
\multicolumn{4}{c}{CityScapes}                     \\ \hline
Method        & $20$ labels     & $50$ labels     & $150$ labels \\ \hline
Supervised    & $50.60$         & $53.90$         & $63.95$          \\
MT            & $48.23$         & $62.14$         & $67.49$          \\
ClassMix      & $56.65$         & $63.89$         & $66.68$      \\\hline
PRCL(w/ClassMix) & $\bm{58.09}$ & $\bm{64.87}$    & $\bm{67.60}$          \\ \hline
\end{tabular}
}
\vspace{-10pt}
\label{tab2}
\end{table}

\noindent\textbf{Results on Pascal VOC.} 
Table \ref{tab1} shows the results of the comparison among our work and baselines on Pascal VOC.
PRCL consistently outperforms other methods at all label rates in our experiment setting.
Notably, our PRCL framework can improve the performance in very limited data situations, where the performance improvement is more meaningful in a semi-supervised setting. 

\noindent\textbf{Results on CityScapes.}
Table \ref{tab2} illustrates the results of comparison among our work and baselines on CityScapes.
PRCL achieves a marginal out-performance over all baselines across a wide range of the number of labeled images.
Similar with the result on Pascal VOC, our framework can unlock more potential in low label rate.
\begin{figure}[t]
  \centering
  \includegraphics[width=0.9\linewidth]{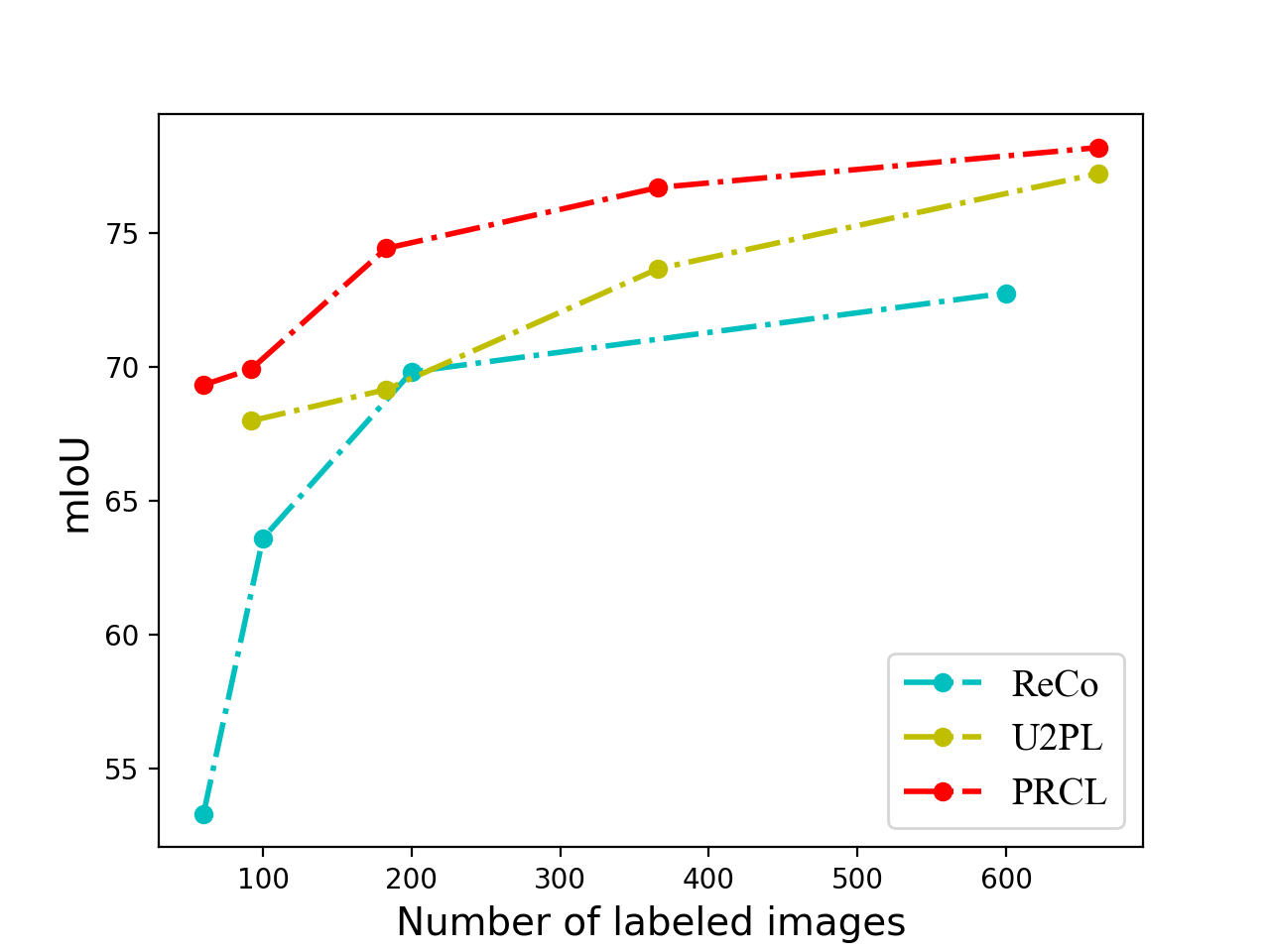}
  \vspace{-5pt}
  \caption{Compare with similar works based on pixel-wise contrastive learning}
  \vspace{-5pt}
  \label{fig4}
\end{figure}
\begin{table}[t]
\centering
\caption{Results on Pascal VOC using a modified ResNet with the deep stem block. The model is trained on the augmented VOC \texttt{train} set, which consists of 10582 samples in total and tested on VOC \texttt{val} set.  And all the results from the recent papers \cite{U2PL,ST++,Reco}.}
\vspace{-7pt}
\setlength{\tabcolsep}{1mm}{%
\begin{tabular}{cccccc}
\hline
\multicolumn{6}{c}{Pascal VOC}                         \\ \hline
Method                 & 92            & 183           & 366          & 662    &\\ \hline
Supervised             & $52.23$       & $62.33$       & $67.11$      & $71.12$  \\ \hline
PseudoSeg(w/ CutOut)   & $57.60$       & $65.50$       & $69.14$      & -     \\
CPS(w/ CutMix)         & $64.07$       & $67.42$       & $71.71$      & $74.48$    \\
PC$^2$Seg(w/ CutOut)   & $57.00$       & $66.28$       & $69.78$      & -     \\ 
U$^2$PL(w/ CutMix)     & $67.98$       & $69.15$       & $73.66$      & $77.21$     \\
ST++(w/ CutOut)        & $65.20$       & $71.00$       & $74.60$      & $74.70$    \\
PSMT                   & $65.80$       & $69.58$       & $76.57$      & -    \\ \hline
PRCL(w/ ClassMix)      & $\bm{69.91}$  & $\bm{74.42}$  & $\bm{76.69}$ & $\bm{77.88}$       \\ \hline
\multicolumn{6}{c}{\textit{Fully-supervised setting (10582 images): 78.20}}   \\ \hline
\end{tabular} %
}
\vspace{-5pt}
\label{tab3}
\end{table}

\noindent\textbf{Comparison with SOTAs.}
We compare our work with following recent SOTAs with their own strong data augmentations (CutOut, CutMix, and ClassMix): PseudoSeg \cite{pseudoseg}, CPS \cite{CPS}, PC$^2$Seg \cite{PC2Seg}, PSMT \cite{psmt}, U$^2$PL \cite{U2PL}, ST++ \cite{ST++}, and ReCo \cite{Reco} on Pascal VOC dataset.
Table \ref{tab3} shows the result.
Our PRCL performs better than other works, especially in the scenarios with low label rates.
Similar works based on pixel-wise contrastive learning \cite{Reco, U2PL} focus on polishing pseudo-labels to pick the correct representation.
Instead, we focus on improving the quality of represetations for robustness under inaccurate pseudo-labels.
The results are shown in Figure \ref{fig4}.
\subsection{Ablation Study}
We conduct the ablation studies based on DeepLabV3+ and ResNet-101 on Pascal VOC.
And all the labeled images are the same as above experiments.

\begin{table}[t]
\centering
\caption{Ablation study on Probabilistic mechanism (P) with different thresholds of sampling strategy.}
\vspace{-7pt}
\setlength{\tabcolsep}{3.7mm}{%
\begin{tabular}{ccccc}
\hline
                          & $\delta_s$& $\delta_w$& 60 labels   & 92 labels \\ \hline
\multicolumn{1}{c}{}      & $0.97$    & $0.70$   & $61.83$     & $66.23$     \\
\multicolumn{1}{c}{w/o P} & $0.90$    & $0.70$   & $60.60$     & $63.02$     \\
\multicolumn{1}{c}{}      & $0.80$    & $0.70$   & $55.45$     & $62.70$     \\ 
                          & $1.00$    & $0.00$   & $59.00$     & $62.92$     \\ \hline
\multicolumn{1}{c}{}      & $0.97$    & $0.70$   & $61.11$     & $65.16$     \\
w/ P                      & $0.90$    & $0.70$   & $61.74$     & $66.19$    \\
                          & $0.80$    & $0.70$   & $\bm{64.27}$& $\bm{67.91}$     \\  
                          & $1.00$    & $0.00$   & $56.78$     & $64.33$     \\ \hline
\end{tabular}%
}
\label{tab4}
\end{table}

\noindent\textbf{Impact of probabilistic representation and distribution prototype.}
To examine the effectiveness of probabilistic mechanism, we compare with deterministic representations and point prototypes. 
And we choose $\ell_2$ distance as a similarity measurement of them.
In addition, to demonstrate the robustness to high-risk false representations, we adjust the strong threshold $\delta_s$ to vary the proportion of unconfident samples.
To amplify the effect of the PRCL loss, we conduct experiments without loss scheduler.
Tabel \ref{tab4} shows the result on 2 label rates.
We observe that under the same settings, our algorithm with probabilistic mechanism performs almost better in both two label rates.
Especially, when lower the strong threshold, our framework will achieve better performance while that with deterministic representation and point prototype is hurt visibly. We mainly attribute it to the more sampled low-confident anchors with low strong threshold.
We argue that there are more critical samples in low-confident predictions since the model does not fully grasp the information of corresponding pixels.
It will be more efficient to concentrate on this part of pixels.
However, low confidence leads to introduce more inaccurate pseudo-labels.
The deterministic representations have no ability to handle these inaccurate pseudo-labels, thus leading to disorder in latent space.
And this dramatically hurts the performance of semantic segmentation.
With our probabilistic mechanism, the model is error-tolerant and reduces the damage to the latent space from inaccurate pseudo-labels while taking full advantage of high-risk but critical representations.
In addition, we conduct the experiments without the sampling strategy ($\delta_s=1$ and $\delta_w=0)$ to demonstrate its necessity.
\begin{table}[t]
\centering
\caption{Ablation study on impact of Loss Scheduler (LS).}
\vspace{-7pt}
\setlength{\tabcolsep}{1.3mm}{%
\begin{tabular}{ccccccc}
\hline
Labels  & 46            & 60             & 92              & 183               & 336                 & 662     \\ \hline
w/o LS  & $\bm{43.31}$  & $\bm{64.27}$   & $67.91$         & $69.40$           & $72.96$             & $75.35$ \\
w/ LS   & $43.00$       & $58.10$        & $\bm{68.52}$    & $\bm{72.14}$      & $\bm{75.05}$        & $\bm{76.12}$      \\ \hline
\end{tabular}%
}
\vspace{-5pt}
\label{tab5}
\end{table}

\noindent\textbf{Impact of Loss scheduler.}
In Table \ref{tab5}, we evaluate the effectiveness of Loss Scheduler $\lambda_c (t)$ (LS).
We can observe that with a high label rate, the model with a loss scheduler performs better than the one without a loss scheduler.
However, with low label rate, model with LS performs slightly worse than model without LS.
We argue that at low label rates, i.e. with less guidance, there will be more inaccurate pseudo-labels.
This case will unlock the potential of probabilistic representations.
Increasing the proportion of contrastive loss in total loss can promote a better distribution in latent space, which will help the downstream task.
When training with a high label rate, pseudo-labels become more reliable.
We should pay more attention to downstream semantic segmentation tasks and decrease the proportion of contrastive loss.
\noindent\textbf{Impact of negative numbers.} We evaluate the performance of different number of negatives used in PRCL framework.
In Table \ref{tab6}, we can observe that performance is better when sampling more negatives. This is because the distribution of sampled negatives is more similar with the true distribution when sample more negatives.
\begin{table}[ht]
\centering
\caption{Ablation study on impact of negative numbers.}
\vspace{-7pt}
\setlength{\tabcolsep}{2.1mm}{%
\begin{tabular}{ccccc}
\hline
Number of negatives & 64 & 128 & 256 & 512   \\ \hline
mIoU                & $65.53$  & $65.72$   & $67.53$   & $\bm{67.91}$ \\ \hline
\end{tabular}%
}
\vspace{-10pt}
\label{tab6}
\end{table}
\section{Conclusion}
In this work, we define pixel-wise representations and prototypes from a new perspective of probability theory and design some tricks to optimize them.
Probabilistic representations alleviate the negative effects from inaccurate pseudo-labels for better performance in contrastive learning.
Besides, a more reliable distribution prototype can be estimated with them.
Based on probabilistic representations and distribution prototypes, we propose a PRCL framework for better representation distribution to improve the performance on semi-supervised semantic segmentation tasks.
Extensive experiments prove the effectiveness of PRCL and its superiority compared with other SOTAs.
As Socrates said, \textit{The only true wisdom is in knowing you know nothing}.
Models need not only what they know, but how much they know.
We hope that this simple but effective framework enables the renaissance of Bayesian theory in computer vision.
\section{Acknowledgements}
This work was supported in part by the Australian Research Council under Project DP210101859 and the University of Sydney Global Development Award (GDA). The authors acknowledge the use of the National Computational Infrastructure (NCI) which is supported by the Australian Government, and accessed through the NCI Adapter Scheme and Sydney Informatics Hub HPC Allocation Scheme.
\bibliography{aaai23}
\end{document}